\documentclass{article}

\usepackage{microtype}
\usepackage{graphicx}
\usepackage{subcaption}
\usepackage{booktabs} %

\usepackage{hyperref}

\usepackage[preprint]{icml2026}

\usepackage{amsmath}
\usepackage{amssymb}
\usepackage{mathtools}
\usepackage{amsthm}
\usepackage{xspace}

\usepackage[capitalize,noabbrev]{cleveref}

\theoremstyle{plain}

\theoremstyle{definition}

\theoremstyle{remark}

\usepackage[textsize=tiny]{todonotes}

\icmltitlerunning{$\mathtt{\mmtldr}$: Unified Evaluation of Information Loss in Multimodal Video Captioning}

\usepackage{cleveref}
\usepackage{listings}
\usepackage[most]{tcolorbox}
\usepackage{enumitem}
\tcbuselibrary{theorems}

\tcbset{
    promptcolor/.style={
        colframe=#1!30!gray,
        colback=#1!10!white,
        colbacktitle=#1!30!gray
    }
}

\newtcbtheorem[
  auto counter,
  number within=section
]{promptbox}{Prompt}{
  promptcolor=gray,
  coltitle=white,
  fonttitle=\bfseries,
  sharp corners=south,
  boxrule=0.8mm,
  width=\textwidth,
  fontupper=\footnotesize,
  breakable,
  before upper={\setlist[itemize]{nosep}}
}{prompt}
\crefname{prompt}{Prompt}{Prompts}

\usepackage{siunitx}
\newcommand{\mmtldr}{ViSIL\xspace}

\begin{document}

\twocolumn[
  \icmltitle{$\mathtt{\mmtldr}$: Unified Evaluation of Information Loss in Multimodal Video Captioning}

  \icmlsetsymbol{equal}{*}

  \begin{icmlauthorlist}
    \icmlauthor{Po-han Li}{equal,ut}
    \icmlauthor{Shenghui Chen}{equal,ut}
    \icmlauthor{Ufuk Topcu}{ut}
    \icmlauthor{Sandeep Chinchali}{ut}
  \end{icmlauthorlist}

  \icmlaffiliation{ut}{The University of Texas at Austin, Texas, USA}

  \icmlcorrespondingauthor{Po-han Li}{pohanli@utexas.edu}
  \icmlcorrespondingauthor{Shenghui Chen}{shenghui.chen@utexas.edu}

  \icmlkeywords{Multimodal Video Summary, Video Captioning, Video-to-Text, Information Theory}
  \vskip 0.3in
]

\printAffiliationsAndNotice{\icmlEqualContribution}

\begin{abstract}
    Multimodal video captioning condenses dense footage into a structured format of keyframes and natural language. By creating a cohesive multimodal summary, this approach anchors generative AI in rich semantic evidence and serves as a lightweight proxy for high-efficiency retrieval.
However, traditional metrics like BLEU or ROUGE fail to quantify information coverage across disparate modalities, such as comparing a paragraph of text to a sequence of keyframes.
To address this, we propose the \textbf{Video Summary Information Loss} (\textbf{\mmtldr}) score, an information-theoretic framework that quantifies the video information not captured by a summary via vision-language model (VLM) inference.
By measuring the information loss, \mmtldr is a unified metric that enables direct comparison across multimodal summary formats despite their structural discrepancies.
Our results demonstrate that \mmtldr scores show a statistically significant correlation with both human and VLM performance on Video Question Answering (VQA) tasks.
\mmtldr also enables summary selection to optimize the trade-off between information loss and processing speed, establishing a Pareto-optimal frontier that outperforms text summaries by $7\%$ in VQA accuracy without increasing processing load.

\end{abstract}

\section{Introduction}
\begin{figure*}[h!]
  \centering
  \includegraphics[width=0.85\textwidth, trim=10 130 10 130, clip]{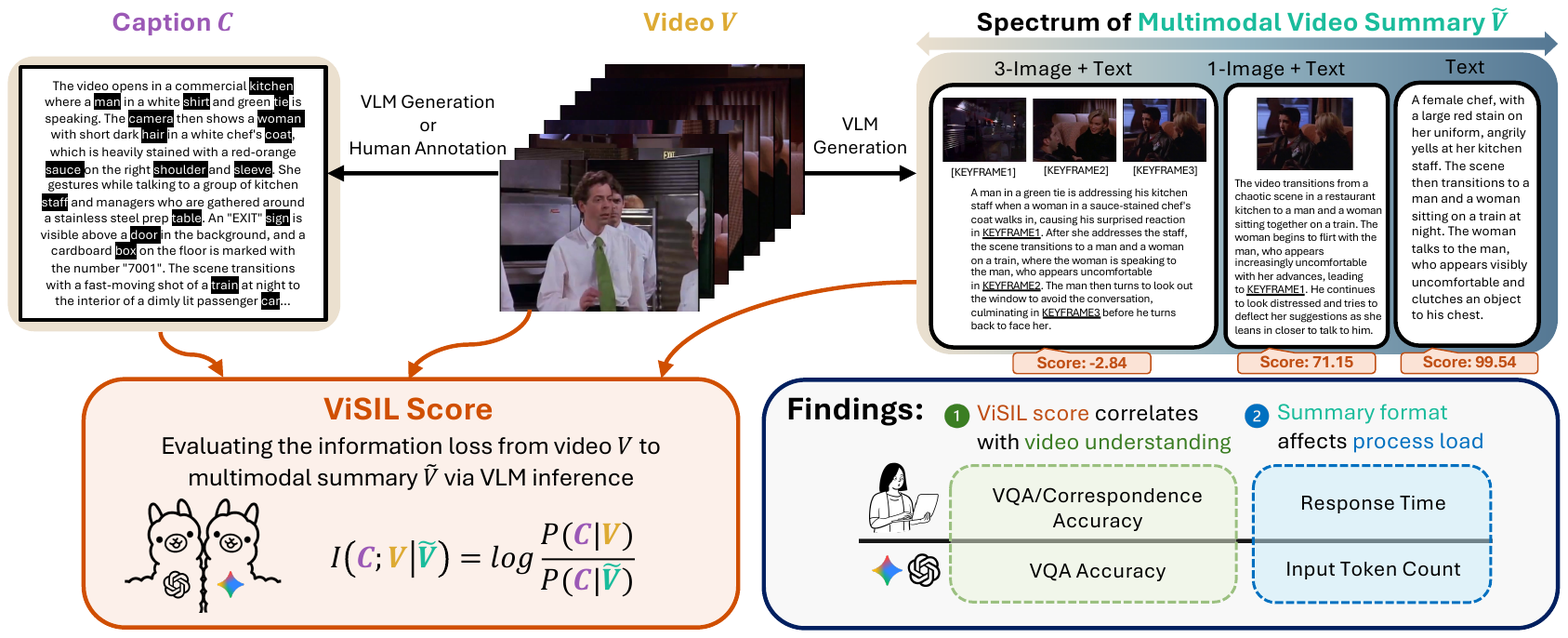}
  \caption{\textbf{A Unified Evaluation for Multimodal Video Captions.}
Given a video $V$, VLM-generated detailed caption $C$, and several multimodal video summaries $\tilde{V}$, the \mmtldr score quantifies the information loss within the summaries relative to the original video content.
Our results show that \mmtldr correlates with video understanding (VQA accuracy) for both humans and VLMs, while the summary format dominates the process load (response time and token count).}
  \label{fig:visil_system}
\end{figure*}

The surge in high-resolution video has rendered multimodal summarization essential. By unifying visual keyframes with linguistic descriptors, multimodal summaries effectively bridge the gap between massive raw datasets and meaningful understanding.
Unlike unimodal descriptions, these multimodal summaries provide the rich semantic grounding required to evaluate text-to-video generation models and the dense indexing necessary for precise retrieval-augmented generation (RAG).
This cross-modal synergy is also critical for human-in-the-loop applications like security surveillance, where combined visual and textual cues enable rapid analysis without reviewing full-length footage. While keyframes capture instantaneous context, text is vital for synthesizing temporal dynamics and providing high-level reasoning that images alone may obscure.
The synergy creates a spectrum of multimodal video summaries, ranging from text-only to hybrid formats with varying keyframe densities, as shown in \cref{fig:visil_system}.

However, this diversity renders traditional evaluation metrics, \textit{e.g.}, BLEU \cite{bleu}, ROUGE \cite{rouge}, or METEOR \cite{banerjee-lavie-2005-meteor}, insufficient for capturing the holistic information contribution across heterogeneous modalities.
These metrics are restricted to unimodal text-to-text comparison and cannot capture information distribution across disparate modalities. Also, it remains unclear which format optimally balances information richness with processing efficiency, such as human response time or the input tokens for a vision-language model (VLM). For instance, it is not yet established whether increasing the number of images in a summary necessarily leads to better video understanding and faster processing.

To unify the evaluation of these heterogeneous formats of modalities, we propose the \textbf{Video Summary Information Loss} (\mmtldr) score, an information-theoretic framework that quantifies the semantic information loss when compressing a video $V$ into a summary $\tilde{V}$.
As illustrated in \cref{fig:visil_system}, we first generate a detailed caption $C$—either through a VLM or human annotation—to act as a comprehensive textual proxy for the source video $V$.
\mmtldr then measures the information loss by evaluating a VLM's ability to recover caption $C$ using the multimodal summary $\tilde{V}$ relative to the original video $V$.
Defined as the conditional pointwise mutual information $I(C; V | \tilde{V}) = \log \frac{P(C | V)}{P(C | \tilde{V})}$, the metric captures visual details that remain ``unaccounted for" by the summary.
By measuring information loss--where lower scores signify better coverage--\mmtldr offers a unified metric that aligns with both human and VLM comprehension across the multimodal summary spectrum.

\textbf{Contributions.}
We introduce \mmtldr, an information-theoretic framework that evaluates diverse summary formats, with human and VLM validation confirming its alignment with Video Question Answering (VQA) performance.
Our work demonstrates that summary format primarily dictates process load—such as response time and token consumption—rather than inherent video understanding.
By leveraging \mmtldr for summary selection, we establish a Pareto-optimal frontier that outperforms pure text summaries by $7\%$ in VQA accuracy without increasing processing overhead.

\section{Related Works}

\textbf{Video Captioning} \cite{qasim2025dense, video_caption_review} is the task of using VLMs to automatically generate a natural language description that semantically summarizes the visual and auditory content of a video. High-quality, precise captions are critical for modern generative AI and data retrieval systems; they are indispensable for semantic grounding in text-to-video generation models \cite{chen2024sharegptvideo, openai_sora2_2025} and crucial for efficient indexing in RAG-based storage and retrieval systems \cite{zhu2023deep, any2any}. While \citet{kudo2023challenging} also explores multimodal (keyframes + text) captioning, they rely on existing unimodal evaluation metrics. In contrast, \mmtldr is for multimodal captions, and we verify it using VLM-based and human-based video understanding tests.

\textbf{Video Caption and Keyframe Evaluation.}
Robust evaluation metrics for high-quality video captions fall into two categories: reference-based methods requiring ground truth captions \cite{kudo2023challenging, chai2025auroracapefficientperformantvideo}, and reference-free methods.
Reference-free methods typically rely on multimodal embedding similarity \cite{lee2020vilbertscore, hessel-etal-2021-clipscore, li2025csa} or mutual information between text and video \cite{chen2025vibe}.
However, embedding-based metrics struggle with cross-format summaries; disparate architectures and incompatible latent spaces—often limited to unimodal text representations \cite{gpt_vector_embedding, gemini_vector_embedding}—preclude a unified metric for heterogeneous data.
In contrast, \mmtldr sidesteps embedding limitations by focusing on cross-modal signal preservation.
This work also departs from prior work, which evaluates modalities in isolation, such as standalone \cite{keyvideollmlargescalevideokeyframe} or VQA-based \cite{Ye_2025_CVPR} keyframe selection.

\textbf{Human-Centric Evaluation.}
Although automatic metrics enable scalable evaluation, human-centric assessment remains the gold standard for measuring the practical utility of video summaries. Prior human evaluations have primarily focused on fluency and informativeness \cite{belz2006comparing, graham2017can}, but these methods are difficult to scale and do not always correlate with practical utility. 
Recent work demonstrates that aligning model representations with human perception--including attention \cite{linsley2018learning}, temporal visual dynamics \cite{parthasarathy2023self}, and conceptual structures \cite{muttenthaler2022human, muttenthaler2025aligning}--improves robustness, interpretability, and generalization.
Extending this philosophy to video summarization, we argue that evaluation should reflect human task performance. We therefore adopt an extrinsic evaluation paradigm \cite{nenkova2011automatic, pu2023summary}, measuring how summaries affect human response time and accuracy on multiple-choice questions grounded in video content.

\section{Video Summary Information Loss (\mmtldr)}

\subsection{Preliminaries}
To establish a theoretical foundation for \mmtldr, we first define Mutual Information (MI) and its pointwise variant.

\textbf{Mutual Information (MI)} \cite{MI} quantifies the mutual dependence between two random variables, $\mathbf{X}$ and $\mathbf{Y}$. It measures how much information is obtained about one random variable through observing the other, thus other literature calls it ``information gain," which is defined as:

\begin{equation}
    \mathbf{I}(\mathbf{X}; \mathbf{Y}) = \mathbb{E}_{X, Y} \left[ \log \frac{P(x, y)}{P(x)P(y)} \right] \geq 0.
\end{equation}

\textbf{Pointwise Mutual Information (PMI)}  \cite{bouma2009normalizedPMI}, in contrast, provides a measure of association between individual events or outcomes $X$ and $Y$, rather than random variables. It is defined as:

\begin{equation}
    \mathcal{I}(X; Y) = \log \frac{P(X, Y)}{P(X)P(Y)} = \log \frac{P(X | Y)}{P(X)} \in (-\infty, \infty).
    \label{eq:pmi}
\end{equation}
MI averages over a distribution and is non-negative; PMI evaluates a single pair and remains unbounded. Since we only use PMI, the notation $\mathcal{I}$ denotes PMI rather than MI throughout this paper.

\subsection{Problem Formulation}
We begin by establishing a mathematical formulation for the problem of multimodal video summarization. Subsequently, we present our proposed \mmtldr score as an approximation for this objective and explain why the approximation is needed for VLMs.

Let $V=I \cup A$ denote a video consisting of $N$ frames $I=\{I_i\}_{i=1}^N$ and an audio track $A$. 
Now, suppose we have a video summary $\tilde{V}=\tilde{I} \cup T$ that consists of a subset of keyframes $\tilde{I} \subseteq I$ and a textual summary $T$.
This formulation provides a flexible definition covering a broad \textbf{spectrum of video summaries}. While $\tilde{V}$ is inherently multimodal, containing both visual and textual components, it seamlessly accommodates unimodal scenarios when either $\tilde{I}=\emptyset$ or $T=\emptyset$.

Now, we aim to evaluate the quality of the video summaries.
We measure the quality of a summary by calculating the PMI as in \Cref{eq:pmi} between the original video and the summary.
A higher PMI score indicates a better summary, as it signifies that the summary contains a greater amount of shared information derived from the video:
\begin{equation}
\begin{aligned}
    \mathcal{I} & (V; \tilde{V}) 
    = \underbrace{\log \frac{P(V | \tilde{V})}{P(V)}}_{\text{(3.1)}} = \underbrace{\log \frac{P(\tilde{V} | V)}{P(\tilde{V})}}_{\text{(3.2)}}
    \label{eq:mi}
    \\ & ~~ (\text{\small PMI of video and summary; higher is better}).
\end{aligned}
\end{equation}

However, direct calculation of the mutual information in \Cref{eq:mi} is doubly intractable.
First, we can only approximate these terms using machine learning models. 
In term (3.1), the numerator is the conditional probability to generate the original video given the summary, $P(V | \tilde{V})$. It is inaccessible by any diffusion models, which are the state-of-the-art video generation models. 
While the denominator, $P(V)$, is the notoriously intractable existential likelihood of a data point. 
The same difficulty holds for the term (3.2) in \Cref{eq:mi} as well.
Therefore, we must approximate the video conditional generation probability and the existential likelihood in \cref{eq:mi}.

\begin{figure*}[h!]
  \centering
  \includegraphics[width=0.75\textwidth, trim=20 150 20 140, clip]{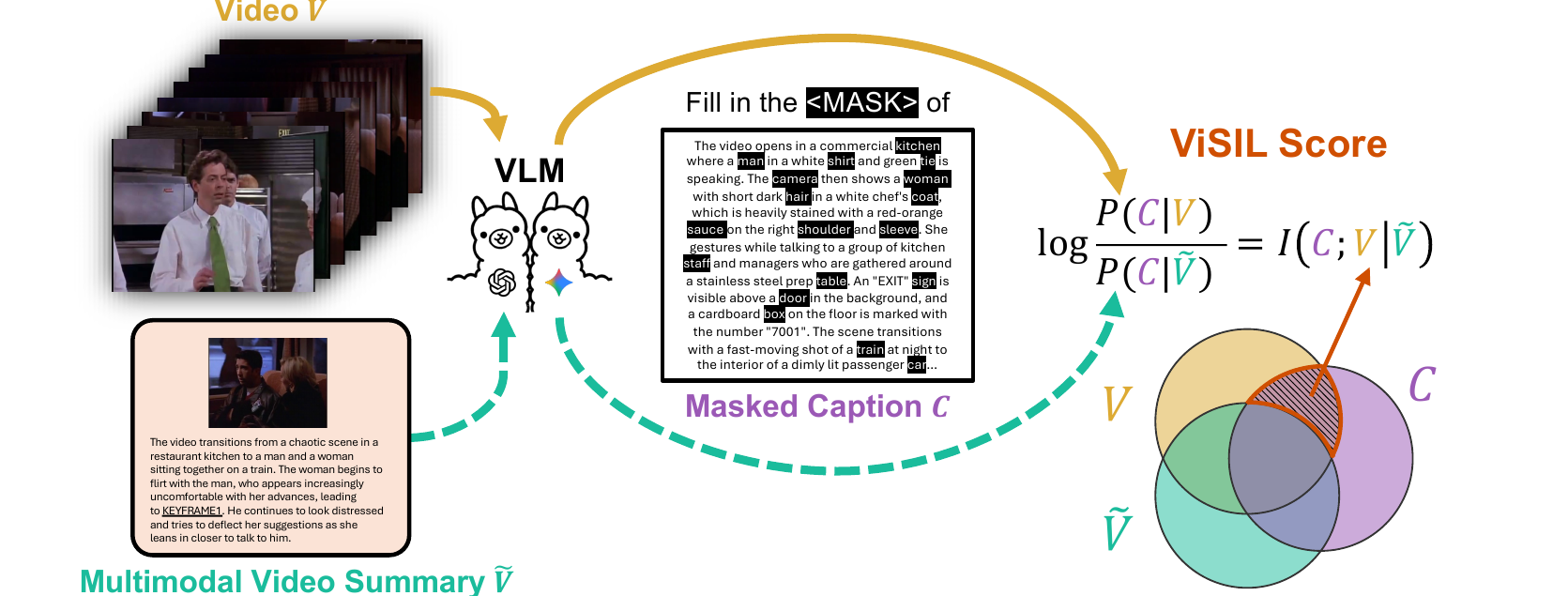}
  \caption{\textbf{ViSIL Implementation via VLM Inference.}
    ViSIL assesses information loss by comparing a VLM's ability to recover masked tokens in caption $C$ from video $V$ versus summary $\tilde{V}$. 
    ViSIL is defined as the pointwise mutual information between the video and caption conditioned on the summary, representing the information in the video that remains unaccounted for by the summary.
    A lower ViSIL score indicates better information preservation.}
  \label{fig:visil_pipeline}
\end{figure*}

\subsection{Approximation with Autoregressive Models}
\label{sec:approximation}

To approximate \Cref{eq:mi} into a solvable form, we draw inspiration from generative models operating in other modalities.
Our key insight is that while video generation models struggle to estimate the necessary conditional probabilities, \textbf{autoregressive VLMs} can leverage their inherent next-token prediction mechanism to estimate conditional probabilities effectively, even when relating different modalities.
The mathematical formulation of VLMs is fundamentally suited, as they model the conditional probability $P_{\text{VLM}}(Y|X)$, which represents the token probabilities of the output sentence $Y$ given the multimodal input $X$.

Based on this insight regarding the solvability provided by next-token mechanisms, we introduce two necessary reformulations: 
First, we require a \textbf{language form of the video} to serve as a textual proxy for the raw video $V$ since VLMs cannot output videos with token probabilities. 
The need for textual reference is common in the evaluation of captions \cite{lee2020vilbertscore, chai2025auroracapefficientperformantvideo}, despite variations in the underlying motivations.
Second, to handle the intractable marginal likelihood term in the denominator, we replace it with a conditional probability of text (\textit{e.g.}, by conditioning it on a prompt or context). This critical transformation converts the difficult marginal $P(\tilde{V})$ into a conditional probability, $P(\tilde{V} | \text{Auxiliary Text})$, which fits perfectly with the next-token prediction mechanism of autoregressive models.

\subsection{\mmtldr--A Unified Framework to Evaluate Video Summaries}
To approximate \Cref{eq:mi}, we first use a VLM to caption the raw video $V$ into a long and detailed caption $C$, which can also be annotated by humans.
Then, we evaluate the information loss between video $V$ and summary $\tilde{V}$---the amount of information contained in the video but missed in the summary, the lower the better.
Using the video caption $C$ as the proxy, we define \mmtldr score as:
\begin{equation}
\begin{aligned}
\mathcal{I}(C; V \mid \tilde{V}) 
&= \log \frac{P(C \mid V, \tilde{V})}{P(C \mid \tilde{V})} 
= \log \frac{P(C \mid V)}{P(C \mid \tilde{V})} 
\label{eq:visil}
\\ & ~~ (\text{\small \mmtldr Score; lower is better}).
\end{aligned}
\end{equation}
The derivation follows from the assumption that $\tilde{V}$ is contained within $V$ by definition (i.e., $\tilde{V} \subseteq V$).
$\mathcal{I}(C; V \mid \tilde{V})$ quantifies how much new information the caption $C$ adds about the raw video $V$, given the multimodal summary $\tilde{V}$. If the multimodal summary already captures all video information, this value should be minimal. For an illustrative example, see \cref{fig:visil_pipeline}.

\cref{eq:visil} defines a unified metric that measures information loss in any multimodal summary $\tilde{V}$ relative to video $V$, enabling direct comparison and selection by minimizing the information loss. 
In our experiments, we demonstrate that the \mmtldr correlates strongly with video understanding tasks assessed by both advanced VLMs and humans.

\textbf{Venn Diagram Interpretation.}
The red shaded area in the Venn diagram (\cref{fig:visil_pipeline}, right) visually represents the information loss $(V\cap C) \setminus \tilde{V}$ that \mmtldr aims to minimize, quantifying the information contained in the video-caption overlap that the summary fails to capture.
By utilizing a comprehensive caption $C$ such that $V \cap C \approx V$ (i.e., $C$ effectively ``covers" $V$), \mmtldr measures how much of the video's core content is missing from the summary $\tilde{V}$.
As the summary becomes more comprehensive, this shaded area shrinks, yielding a lower (better) score.

To link \cref{eq:mi} to \cref{eq:visil}, note that maximizing the overlap $V \cap \tilde{V}$ (as in \cref{eq:mi}, shown by the yellow-green intersection) is equivalent to minimizing the set difference $V \setminus \tilde{V}$.
When the caption is sufficiently descriptive ($V \cap C$ is large enough), the shaded region measured by \cref{eq:visil} essentially measures $V\setminus \tilde{V}$.

\textbf{Mitigating Hallucination.}
Inevitably, VLMs may hallucinate during generation. To mitigate hallucinations, \mmtldr employs distinct models for generation and evaluation, so the evaluation is less biased and the hallucinated content is not reinforced.
\mmtldr inherently minimizes hallucination impact, as shown in the Venn Diagram: if caption $C$ contains ungrounded content, the score remains robust as $I(V;C)$ filters out information not present in the source video $V$.  Similarly, hallucinations in summary $\tilde{V}$ do not assist in recovering grounded tokens in $C$. Because \mmtldr approximates the reduction in uncertainty of $V$ given $\tilde{V}$, such hallucinations act as noise that fails to decrease information loss, effectively penalizing ungrounded summaries.

\begin{figure}
    \centering
    \includegraphics[width=0.85\linewidth]{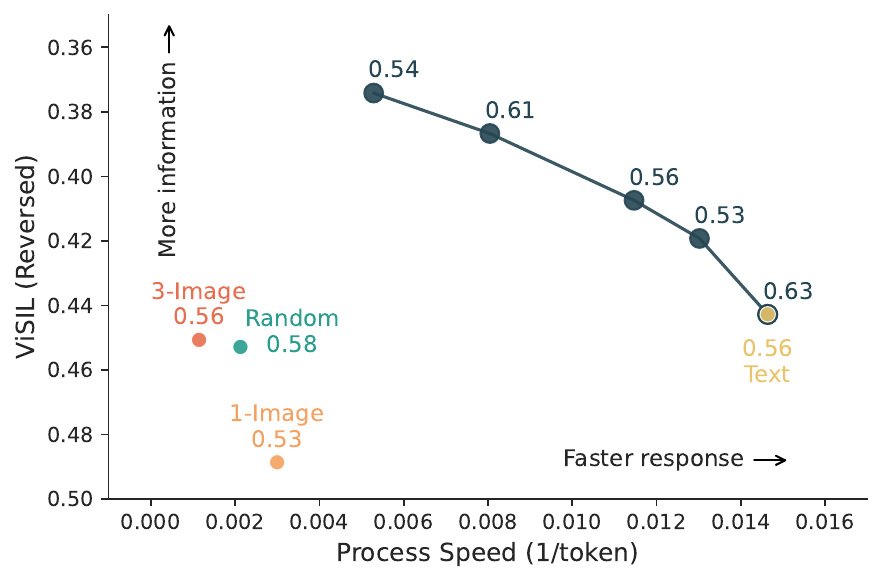}
    \caption{\textbf{Pareto Frontier of Process Speed and \mmtldr Score} showing that static formats are sub-optimal for the process speed–information trade-off.
    \textit{The annotated VQA accuracy} confirms \mmtldr identifies high-utility summaries that outperform pure text and fixed-image formats while preserving processing speed.}
    \label{fig:pareto}
\end{figure}

\begin{table*}[t!]
    \centering
    \begin{tabular}{l S[table-format=2.2(2)] S[table-format=3.2(2)] S[table-format=3.2(2)] S[table-format=5.2(7)]}
    \toprule
    \textbf{Metric} & {\textbf{Text Only}} & {\textbf{1-Image}} & {\textbf{3-Image}} & {\textbf{Video}} \\ \midrule
    MVBench (\texttt{EpR}) (token) & 59.05 \pm 8.58 & 326.34 \pm 10.84 & 866.88 \pm 11.04 & 7148.02 \pm 2348.49 \\
    LongVideoBench (\texttt{SSS}) (token) & 77.86 \pm 11.68 & 336.98 \pm 10.76 & 873.27 \pm 11.57 & 54587.73 \pm 54131.49 \\ 
    \midrule
    Human Response Time (sec) & 62.60 \pm 35.98 & 64.78 \pm 41.61 & 65.94 \pm 35.68 & 85.23 \pm 78.57 \\
    \bottomrule
    \end{tabular}
    \vspace{0.5em}
    \caption{\textbf{Comparison of VLM and Human Process Load across Summary Formats.} Video incurs the highest process load among all formats, while the text-only format yields the lowest.}
    \label{tab:efficiency_metrics}
\end{table*}

\begin{figure*}[t!]
    \centering
    \begin{minipage}{0.45\textwidth}
        \centering
        \includegraphics[width=\linewidth]{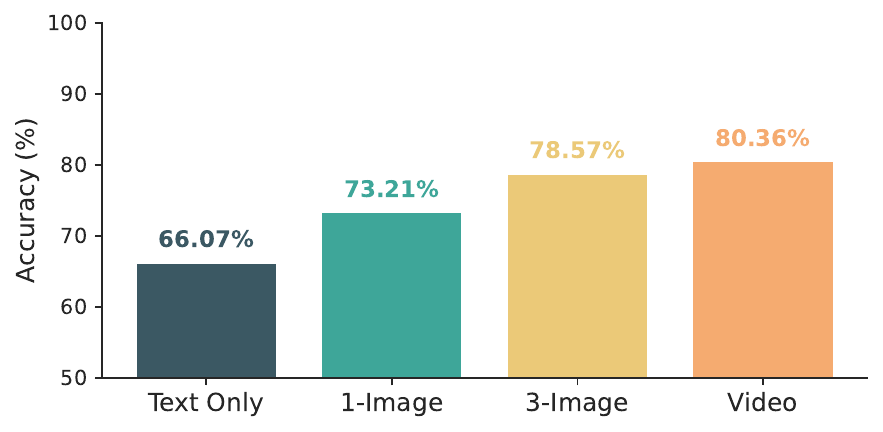}
        \subcaption{VQA Accuracy}
        \label{fig:human_acc}
    \end{minipage}
    \hfill
    \begin{minipage}{0.45\textwidth}
        \centering
        \includegraphics[width=\linewidth]{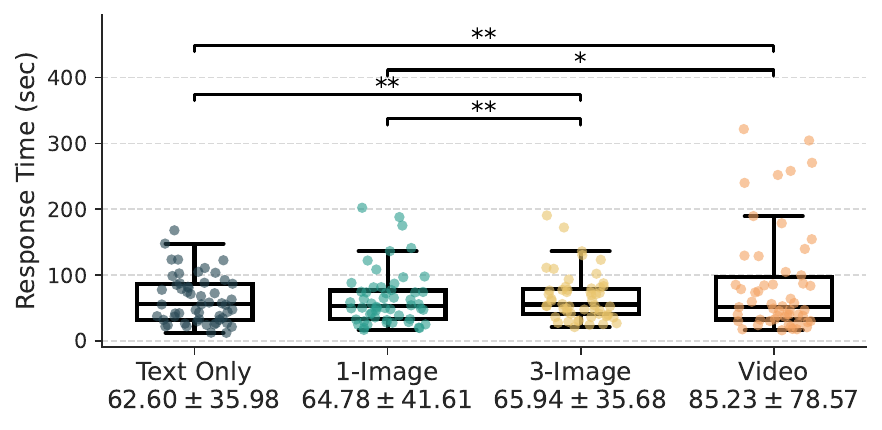}
        \subcaption{Response Time Distribution}
        \label{fig:human_rt}
    \end{minipage}
    \caption{\textbf{Human Performance across Summary Formats.} (a) Accuracy improves as visual context increases, with the 3-Image format approaching the ceiling set by Video. (b) Response times remain stable across all static formats but increase significantly for Video.
    $^*$ denotes statistical significance at $p < 0.05$; $^{**}$ denotes $p < 0.01$.}
    \label{fig:human_performance}
\end{figure*}

\subsection{Information Loss-Process Efficiency Trade-off}
We propose a \mmtldr-based summary selection strategy to balance information loss and processing load by minimizing the joint objective:
\begin{equation}
    \min_{\tilde{V}} \quad \mathcal{I}(C; V | \tilde{V}) + \alpha \cdot \tau(\tilde{V}),
    \label{eq:optimization}
\end{equation}
where $\tau(\tilde{V})$ denotes the token count (processing load), and $\alpha$ is the Lagrange multiplier controlling the trade-off.
Varying $\alpha$ traverses the Pareto-optimal frontier, selecting summaries that balance semantic completeness with the processing speed of VLMs or humans. As shown in \cref{fig:pareto}, \mmtldr dominates fixed baselines like \textit{Random} (randomly selected summaries), \textit{1-Image}, and \textit{3-Image} summaries. The setup is detailed later in \cref{sec:exp}. Notably, at the highest processing speeds, our selected summaries achieve $63\%$ accuracy, significantly outperforming the $56\%$ accuracy of pure text summaries at equivalent efficiency.
\cref{fig:pareto} confirms that our optimization effectively preserves understanding utility while minimizing overhead.

\subsection{\mmtldr Computation via VLM Inference}
To compute \mmtldr, we only require next-token probabilities from VLMs, which all major APIs and local models provide. Since most online APIs produce non-deterministic probabilities \cite{he2025nondeterminism}, we estimate stable values by repeated sampling and taking the geometric mean.

To accelerate evaluation, we follow prior work \cite{chen2025vibe,jung2024informationtheoretic, uncap} and approximate sentence probability using keyword prediction. We mask key semantic tokens and compute
\[P(C|V) \simeq \prod_{i=1}^n P(k_i|V),\quad P(C|\tilde{V)} \simeq \prod_{i=1}^n P(k_i|\tilde{V}),\]
where $k_i$ denotes keywords in $C$. Keyword-based estimation reduces VLM inference cost, improves stability, and ignores low-information tokens such as `a', `the', etc.

\section{Experiment and User Study}
\label{sec:exp}

Let $\mathcal{D} = \{(V_i, Q_i, A_i)\}_{i=1}^M$ denote a visual question-answering (VQA) dataset where each sample contains a video $V$, a question $Q$, and an answer $A$.
Our evaluation focuses on two specialized subsets: Episodic Reasoning (EpR) from MVBench \cite{li2024mvbench}, which tests long-term temporal understanding, and the SSS (Sequence of Scenes) subset of LongVideoBench \cite{wu2024longvideobench}.
For each video $V$ in the dataset, we generate a detailed caption $C$ using \texttt{Gemini 2.5 Pro}, as \texttt{Gemini 3} came out after we conducted the experiments.
Recall that the multimodal summary $\tilde{V} = \tilde{I} \cup T$ consists of keyframes $\tilde{I}$ and text $T$. We constrain the number of keyframes in $\tilde{I} = \{f_1, \dots, f_k\}$ to $k \leq 3$.
See dataset selection details and all prompts used in Appendices~\ref{app:dataset_selection} and \ref{app:prompt}.

\begin{figure*}[t!]
    \centering
    \begin{subfigure}{0.32\textwidth}
        \centering
        \includegraphics[width=\textwidth]{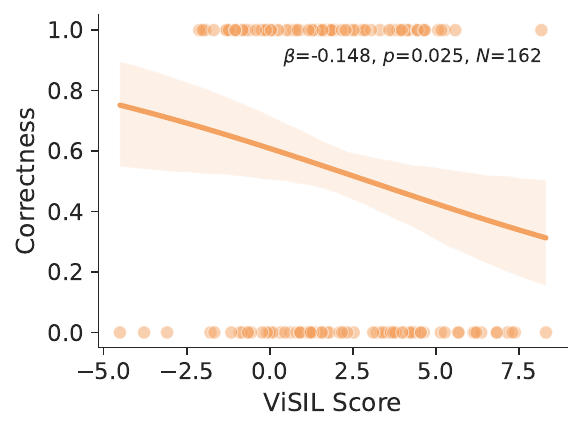}
        \caption{MVBench (\texttt{EpR})}
        \label{fig:mvbench}
    \end{subfigure}
    \begin{subfigure}{0.32\textwidth}
        \centering
        \includegraphics[width=\textwidth]{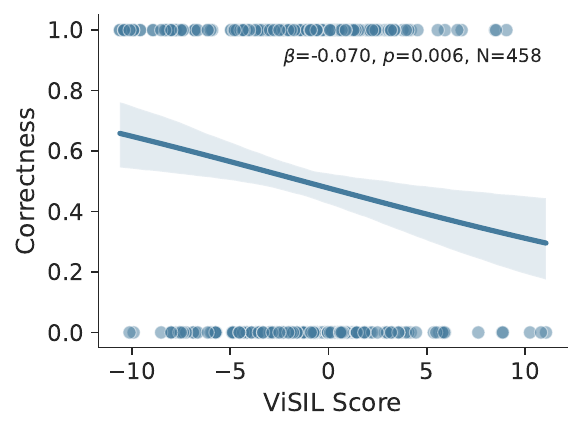}
        \caption{LongVideoBench (\texttt{SSS})}
        \label{fig:longvideobench}
    \end{subfigure}
    \begin{subfigure}{0.32\textwidth}
        \centering
        \includegraphics[width=\textwidth]{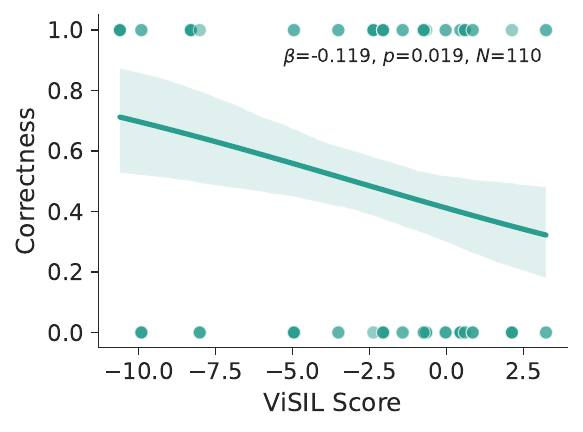}
        \caption{Human VQA}
        \label{fig:human}
    \end{subfigure}

    \caption{\textbf{Logistic regression analysis} between individual sample correctness and corresponding \mmtldr scores. The downward trend in general indicates that higher \mmtldr scores (representing more information loss) correlate with decreased VLM accuracy.}
    \label{fig:rq1_corr}
\end{figure*}

\textbf{Captioning \& Keyword Masking $C$.} 
We employ a two-stage pipeline. First, \texttt{Gemini 2.5 Pro} generates comprehensive captions capturing events and legible text. Second, \texttt{GPT-5} extracts up to $20$ fine-grained keywords, preserving their original morphology and sequential order.

\textbf{Summary Construction $\tilde{V}$.}
We use \texttt{Gemini 2.5 Pro} to both select representative keyframes $\tilde{I}$ and generate the textual component $T$. Crucially, $T$ is conditioned on $\tilde{I}$ to ensure the summary is visually grounded and contextually faithful to the underlying content. The final summary $\tilde{V}$ is thus a composite of the generated text description and the retrieved keyframes.

\subsection{Research Questions}
We investigate the role of multimodal video summaries in supporting both VLM and human understanding of video content. We address the following research questions (RQ):

\begin{enumerate}[label=\textbf{RQ\arabic*.}, leftmargin=*]
    \item \textbf{\mmtldr as a predictive metric.} To what extent does the \mmtldr score correlate with downstream VLM and human video understanding, as measured by performance on video question answering tasks?
    
    \item \textbf{Impact of summary format.} How do different summary formats (\textit{Text-Only}, \textit{1-Image}, \textit{3-Image}, and \textit{Full Video}) affect comprehension performance for both VLMs and human users?
\end{enumerate}

\subsection{VLM Evaluation}
We first compute the \mmtldr score using \texttt{Gemini 2.0 Flash}, as specified in Appendix~\ref{prompt:visil_score_compute}. Then, we also evaluate VLM performance on the VQA task, where we employ \texttt{Gemini 2.5 Pro} as the answering model, following the evaluation protocol detailed in Appendix~\ref{prompt:vqa_eval}. 

For each video and each multimodal summary, we sample $3$ independent runs and take the geometric average of token probabilities to account for generation variability.
For each VQA question, the model is provided only with the corresponding summary and is tasked with answering the associated question. We then compute (1) the VQA accuracy achieved under each summary format and (2) the \mmtldr score of the same summary, enabling a paired analysis at the instance level.

\begin{figure*}[h]
    \centering
    \includegraphics[width=0.85\textwidth]{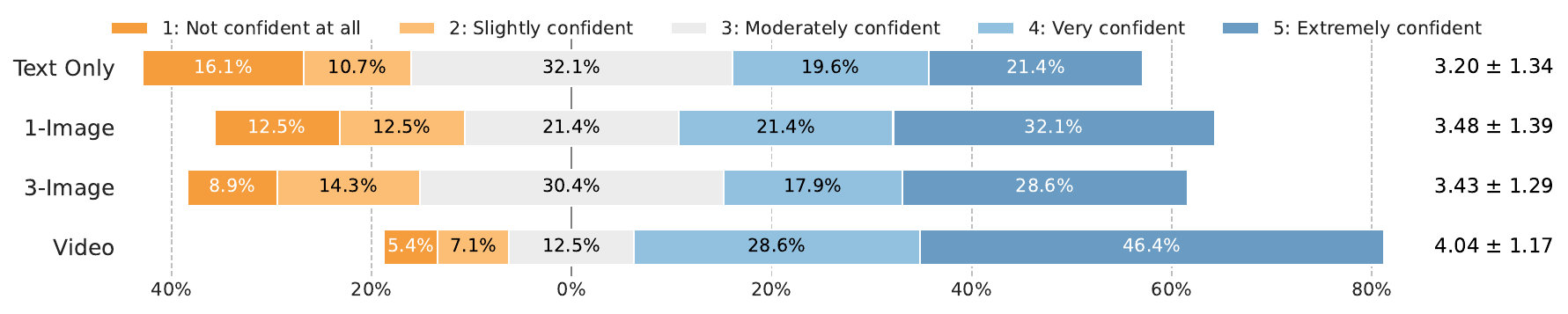}
    \caption{\textbf{Distribution of Human Confidence Ratings.} A diverging stacked bar chart of Likert responses (1--5) shows that dynamic visual context (Video) shifts the distribution toward higher confidence compared to static summaries.}
    \label{fig:human_confidence}
\end{figure*}

\begin{figure*}[t!]
    \centering
    \includegraphics[width=0.4\linewidth, trim=0 25 0 25, clip]{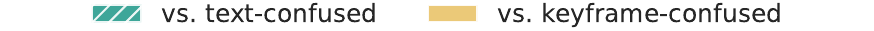}
    
    \begin{subfigure}[b]{0.43\textwidth}
        \centering
        \includegraphics[width=\linewidth]{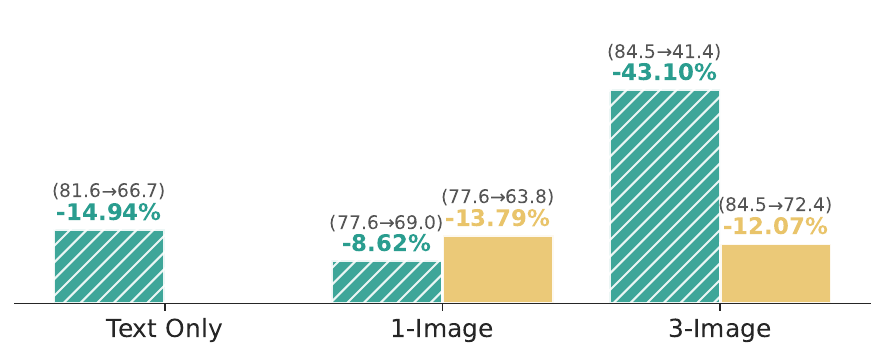}
        \caption{Accuracy Drop (\%)}
        \label{fig:corr_acc}
    \end{subfigure}
    \hfill
    \begin{subfigure}[b]{0.43\textwidth}
        \centering
        \includegraphics[width=\linewidth]{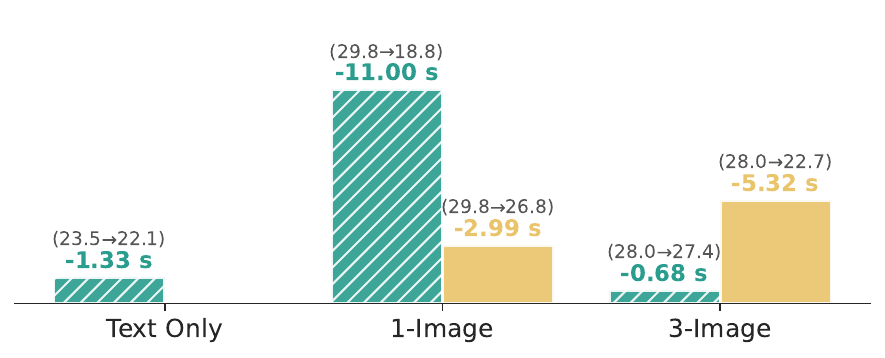}
        \caption{Decrease in Response Time (s)}
        \label{fig:corr_rt}
    \end{subfigure}
    
    \caption{\textbf{Sensitivity to Information Inconsistency.} (a) Accuracy drops show stable sensitivity to visual confusion (yellow), while sensitivity to textual errors (green) diminishes in 3-Image format.
    (b) Response times mirror this trend, with users identifying errors faster in 1-Image summaries than with 3-Image format.}
    \label{fig:correspondence_results}
\end{figure*}

Our evaluation reports two metrics: (1) VQA accuracy per summary format and (2) correlation between \mmtldr scores and VQA accuracy across formats and videos. This analysis allows us to quantify how well \mmtldr reflects downstream task performance and to assess its effectiveness as a proxy for summary informativeness in VLM-based reasoning.

\subsection{User Study}
We conducted two controlled user studies to evaluate human understanding of video summaries. Both utilized a within-subjects balanced Latin square design \cite{keedwell2015latin} to mitigate ordering effects. Detailed demographics and interfaces are provided in Appendices \ref{app:demograph} and \ref{app:study_ui}.
To isolate human recall from active information retrieval—such as re-watching or scrubbing through footage—we restrict users to a single-pass viewing. It ensures that response times measure immediate comprehension rather than the latency involved in searching for specific details.
This single-viewing approach better captures the essence of video understanding by measuring human information acquisition rather than the navigational efficiency of video.

\textbf{VQA Test.}
We measured \textit{accuracy}, \textit{response time}, and \textit{confidence} (5-point Likert \cite{likert1932technique}) across four summary conditions: \textit{Text-Only}, \textit{1-Image}, \textit{3-Image}, and \textit{Full Video}. Participants ($N=37$) answered questions for 4 unique videos from LongVideoBench, seeing each video in only one format to prevent learning effects.

\textbf{Correspondence Test.}
While VQA tests video understanding, it does not verify if users can detect ungrounded content. To investigate human sensitivity to information inconsistencies (ungrounded content), we conducted a correspondence test across $3$ formats: \textit{Text-Only}, \textit{1-Image}, and \textit{3-Image}.  
Summaries were either \textit{ground truth} (original) or \textit{confused} (adversarial distractors generated by perturbing text or keyframes via \texttt{GPT-5-Chat}). Using $6$ unique videos ($3$ per dataset), participants ($N=29$) were first shown the original video and instructed to identify whether subsequent summaries correctly matched the video as quickly as possible.

\section{Results and Analysis}

The experimental results show that \mmtldr reliably predicts both human and VLM video understanding, while summary format determines process efficiency and human sensitivity to information inconsistency.

\subsection{RQ1: \mmtldr as a Predictor of Video Understanding}
To understand how visual information loss affects video understanding, we examine the relationship between \mmtldr scores and task correctness using logistic regression since correct and incorrect are binary.
As illustrated in \Cref{fig:rq1_corr}, our evaluation reveals a consistent, statistically significant negative correlation between \mmtldr scores and correctness. 

On MVBench, higher information loss (higher \mmtldr) effectively predicts lower VQA accuracy ($\beta = -0.148, p = 0.025, N = 162$). 
This predictive power extends to LongVideoBench as well, where we observe a significant negative correlation ($\beta = -0.070, p = 0.006, N = 458$), indicating that information density is a critical factor even for long-form video understanding.

This trend is further validated by human study, which closely mirrors the VLM results: human VQA correctness exhibits a significant negative correlation with \mmtldr scores ($\beta = -0.119, p = 0.019, N = 110$). This alignment suggests that \mmtldr captures an intrinsic information loss of summaries. Rather than merely reflecting model-specific biases, the metric quantifies a fundamental loss of semantic utility in condensed video representations.

To further validate the robustness of these correlations, we conducted a model-agnostic permutation test \cite{moore1999bootstrapping, fisher1971design, ptman1937significance} on Pearson’s coefficient. As detailed in Appendix~\ref{app:permu_test}, the results maintain statistical significance across all datasets, confirming that the observed inverse relationship between information loss and performance is not a product of random chance.

\textbf{Video Understanding and VQA.}
While VQA is a critical task, it typically focuses on specific subsequences or frames, failing to capture the full semantic scope of a video.
In contrast, \mmtldr provides a holistic measure of information loss between a video and its summary, addressing a broader conceptual requirement than targeted question-answering. 
Thus, while \mmtldr scores demonstrate a strong correlation with VQA accuracy, they represent a distinct metric for overall video understanding rather than a direct equivalent.

\begin{figure}
    \centering
    \includegraphics[width=0.85\linewidth]{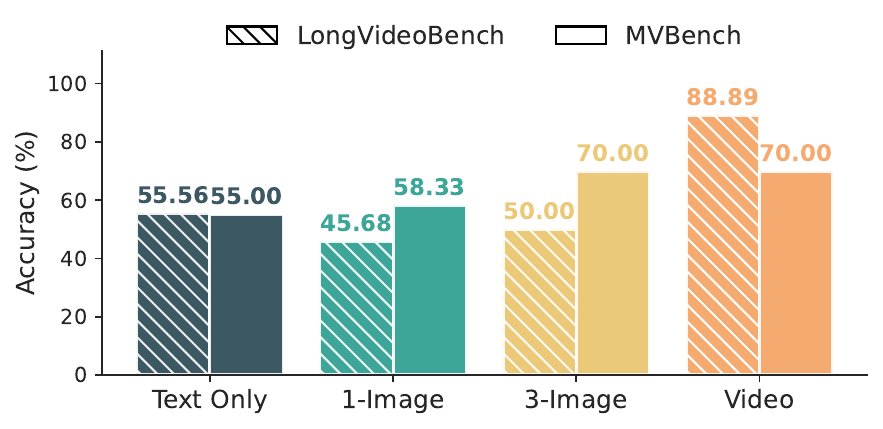}
    \caption{\textbf{VLM VQA accuracy across varying summary formats} improves as the number of visual keyframes increases.}
    \label{fig:rq2_vlm_acc}
\end{figure}

\subsection{RQ2: Impact of Summary Format}

\textbf{On Process Load.}
\Cref{tab:efficiency_metrics} highlights the variance in computational and cognitive costs. 
While full video inputs provide maximum context, they incur a prohibitive computational overhead, consuming up to $700\times$ more VLM tokens than text-only summaries ($54.5\text{k}$ vs.\ $77$ tokens on LongVideoBench) and approximately $62\times$ more than the 3-Image summary ($873$ tokens).
Human cognitive load follows a similar trend; participants required significantly more time to process full video ($85.23$s) compared to static summaries ($\approx 64$s). 
Notably, the 3-Image format incurs only a marginal increase in response time over text ($+3.34$s) while maintaining a relatively low token footprint, suggesting it offers an efficient middle ground.

\textbf{On Human Performance.}
We analyze how summary formats impact human performance and subjective certainty. \Cref{fig:human_performance} and \ref{fig:human_confidence} illustrate the trade-off between accuracy, efficiency, and confidence.
As expected, the full Video format achieves the highest VQA accuracy ($80.36\%$). However, the 3-Image summary is remarkably competitive, achieving $78.57\%$ accuracy--within $2\%$ of the full video baseline--while reducing human response time by nearly $20$ seconds on average ($65.94$s vs.\ $85.23$s). It identifies that the 3-Image format preserves semantic information without the temporal redundancy of video.
We observe a similar trend in VLM performance (\cref{fig:rq2_vlm_acc}), where increasing visual density enhances grounding. 

While performance is similar, user perception differs. As shown in \Cref{fig:human_confidence}, participants reported significantly higher confidence when viewing full videos ($4.04 \pm 1.17$) compared to the 3-Image format ($3.43 \pm 1.29$). This suggests that while static summaries are sufficient for \textit{correct} reasoning, the dynamic context of video provides a psychological layer of reassurance that static keyframes lack.

\textbf{Sensitivity to Information Inconsistency.}
We then pivot to human evaluation to investigate the robustness of content understanding, \textit{i.e.}, whether humans remain \textbf{sensitive} to information inconsistencies (\Cref{fig:correspondence_results}). In our correspondence tests, participants were asked to verify if a summary accurately represented a video they had just viewed. Our results demonstrate that humans are sensitive to both textual and visual perturbations:
\begin{itemize}
    \item \text{Sensitivity to \textit{textual} confusion:} Human sensitivity to textual hallucinations is fragile. While the 1-Image format preserves sensitivity (suffering only a $-8.62\%$ accuracy drop when text is perturbed), the 3-Image format significantly \textbf{dampens sensitivity}. In the 3-Image condition, accuracy plummets by $43.10\%$, indicating that the increased visual context masks textual errors, causing users to overlook them.

    \item \text{Sensitivity to \textit{visual} confusion:} In contrast, sensitivity to visual inconsistencies remains stable; swapping keyframes results in a comparable accuracy decrease for both 1-image ($-13.79\%$) and 3-image ($-12.07\%$) formats, suggesting that users maintain a consistent baseline of visual attention regardless of image count.
\end{itemize}

We further validated this setup using an \textbf{LLM-as-Judge} on the same correspondence test with the same human prompt. Despite high baseline accuracy ($100\%$ on MVBench; $77.8\%$ on LongVideoBench), \texttt{Gemini 2.5 Pro} failed to detect visual inconsistencies, with accuracy plummeting to $50\%$ in both \textit{1-Image} and \textit{3-Image} formats. Conversely, the model remained robust to textual perturbations.  
This finding aligns with existing observations that VLMs are more susceptible to visual confusion than textual inconsistencies, likely due to the dominance of textual data in their pre-training corpora.

\subsection{Discussion and Limitations}
\mmtldr measures cross-modal information retention rather than text overlap, so the experiments exclude BLEU-style metrics that fail in cross-modal evaluation. 
Although \mmtldr remains stable across VLM backbones, scores remain incomparable across models due to model-specific bias. Following \citet{chen2025vibe}, we exploit the lower complexity of evaluation relative to captioning, enabling small and efficient \texttt{Flash} VLMs to act as reliable evaluators.

Despite its advantages, \mmtldr has limitations. The method could evaluate audio-integrated summaries if a VLM supports audio next-token prediction, but we omit such experiments due to summary generation complexity. The metric also depends on base-model multimodal strength, since \mmtldr requires a textual proxy $C$ from video captioning; we mitigate this reliance by using strong \texttt{Gemini} models. Finally, \mmtldr targets evaluation only and does not address summary generation or keyframe selection, which forms an NP-hard problem \cite{sun2025mdp3}.

\section{Conclusion and Future Works}
We introduce \mmtldr, an information-theoretic framework for unified evaluation of multimodal video summaries. Unlike traditional metrics limited to specific modalities, \mmtldr quantifies the video information loss across any summary format.
\mmtldr captures information richness grounded in the source video, while the summary format determines processing load (\textit{e.g.}, human response time, VLM tokens). Thus, \mmtldr serves as a proxy for information density, separating information content from processing efficiency.

Future work can expand \mmtldr to handle interactive summarization where the ``information need" may shift based on user queries. 
It can also include audio in the spectrum of video summaries while still leveraging the \mmtldr framework. 
We aim to explore the integration of \mmtldr into the training loop of video summarization models to directly optimize for information preservation rather than just evaluation.

\clearpage
\section*{Impact Statement}
This work presents a novel evaluation metric, \mmtldr, designed to advance the field of multimodal video summary for video understanding and captioning.
By providing a unified method to assess the quality of both textual and visual summaries, our research facilitates the development of more accurate AI systems that can improve accessibility for the visually impaired and optimize large-scale video retrieval.
We acknowledge that advancements in automated video analysis carry inherent ethical risks regarding privacy and potential surveillance; therefore, we emphasize the use of such metrics for enhancing information transparency and user utility.
To ensure the reliability of our metric, we conducted a human study to align our mathematical formulation with human judgment. 

This study was performed with Institutional Review Board (IRB) approval and strict adherence to ethical standards, ensuring participant anonymity and the responsible handling of data. We believe the societal consequences of this work are positive and do not feel any specific negative impacts must be highlighted beyond these general considerations.

\bibliography{bibtex/reference}
\bibliographystyle{icml2026}

\newpage
\appendix
\onecolumn
\section*{Appendix}

\section{Dataset Selection}
\label{app:dataset_selection}

To evaluate the capabilities of our framework in long-context video understanding and reasoning, we utilize two challenging benchmarks: LongVideoBench \cite{wu2024longvideobench} and MVBench \cite{li2024mvbench}. We specifically select tasks that necessitate sustained temporal attention and high-level semantic synthesis rather than simple object recognition. 

From LongVideoBench, we focus on the \texttt{Sequence of Scenes (SSS)} task, which requires identifying the correct chronological order or relationship between disparate events. From MVBench, we utilize the \texttt{Episodic Reasoning (EpR)} task, which tests a model's ability to infer causal links and overarching narratives across extended durations. The details of these subsets are summarized in \Cref{tab:dataset_stats}.
In LongVideoBench, Gemini refuses to generate one 3-Image video summary due to sensitive content after several retries, while still generating summaries in other formats of the same video.

\begin{table}[h]
\centering
\caption{Statistics of Selected Dataset Subsets }
\label{tab:dataset_stats}
\begin{tabular}{@{}lccc@{}}
\toprule
\textbf{Task Subset} & \textbf{Videos} & \textbf{QA Pairs} & \textbf{Avg. Duration (sec)} \\ \midrule
LongVideoBench (\texttt{SSS}) & 37 & 54 & $207.52\pm203.16$ \\
MVBench (\texttt{EpR}) & 20 & 20 & $39.81\pm13.66$ \\ \bottomrule
\end{tabular}
\end{table}

\section{\mmtldr Distribution Scatters}
\label{app:visil_dist}

\begin{figure}[h]
    \centering
    \begin{subfigure}[b]{0.45\linewidth}
        \centering
        \includegraphics[width=\linewidth]{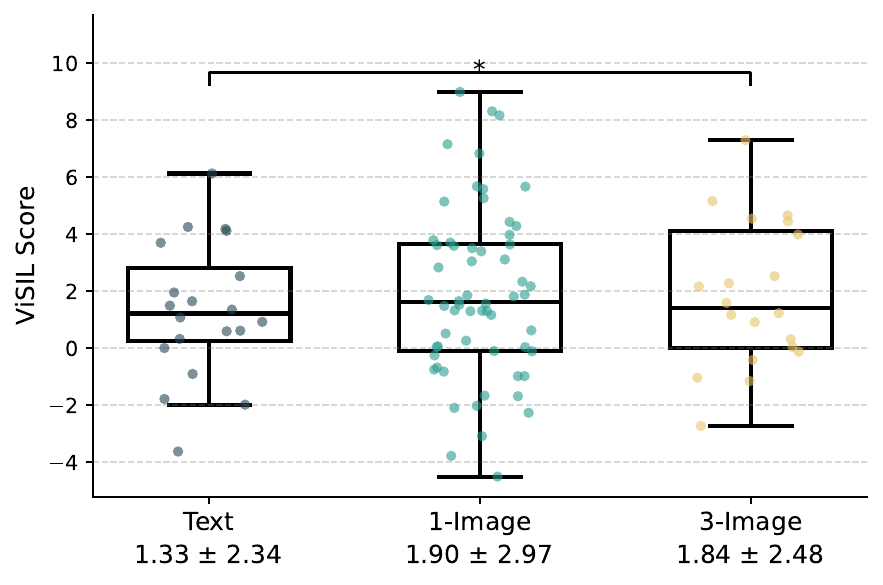}
        \caption{MVBench}
        \label{fig:visil_dist_mvbench}
    \end{subfigure}
    \begin{subfigure}[b]{0.45\linewidth}
        \centering
        \includegraphics[width=\linewidth]{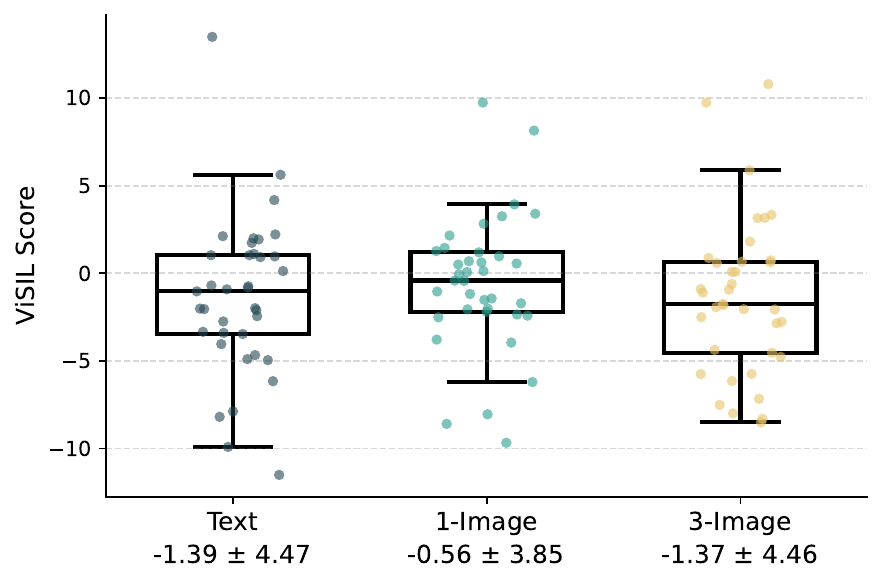}
        \caption{LongVideoBench}
        \label{fig:visil_dist_longvideobench}
    \end{subfigure}
    
    \caption{\textbf{Distribution of \mmtldr Scores.} We visualize the scatter of \mmtldr scores on (a) MVBench and (b) LongVideoBench.}
    \label{fig:visil_dist}
\end{figure}

As shown in \Cref{fig:visil_dist}, we analyze the distribution of \mmtldr scores to verify the metric's discriminative power across different datasets. On both MVBench and LongVideoBench, the scores exhibit consistent distributions and overlapping quartiles across Text, 1-Image, and 3-Image formats. The comparable mean scores suggest that \mmtldr is modality-agnostic, evaluating the intrinsic informativeness of a summary rather than its specific format.

\section{Participant Recruitment and Demographics}
\label{app:demograph}
We conducted a Video Question Answering (VQA) test and a correspondence test. Participants were recruited via Prolific \cite{palan2018prolific}. All were adults (age $>18$) with normal or corrected-to-normal vision and English proficiency. Demographic details are summarized in \Cref{tab:demographics}.
\newpage

\begin{table}[ht!]
\centering
\caption{Participant Demographics}
\label{tab:demographics}
\begin{tabular}{@{}lcccc@{}}
\toprule
\textbf{Study} & \textbf{$N$} & \textbf{Mean Age ($SD$)} & \textbf{Gender Distribution} \\ \midrule
VQA Test            & 37 & 34.32 (11.76) & 64.9\% Male, 35.1\% Female \\
Correspondence Test & 29 & 34.21 (11.97) & 79.3\% Male, 17.2\% Female, 3.4\% Non-binary \\ \bottomrule
\end{tabular}
\end{table}

\section{User Study Instruction and Interfaces}
\label{app:study_ui}
\begin{figure*}[h!]
    \centering
    \begin{subfigure}{0.45\textwidth}
        \centering
        \includegraphics[width=\textwidth]{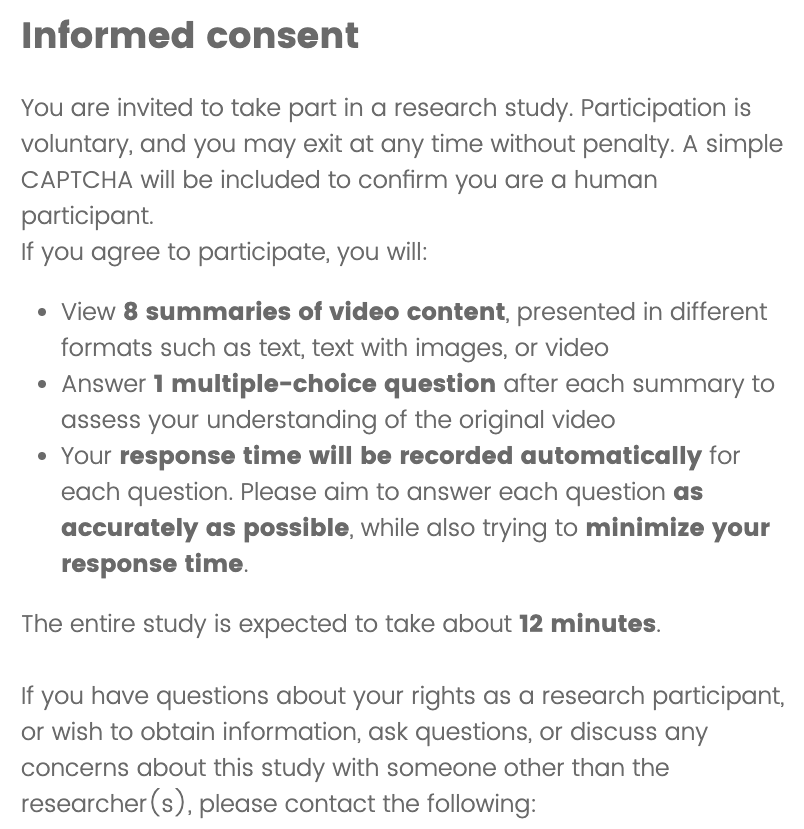}
        \caption{Instructions for VQA test.}
    \end{subfigure}
    \begin{subfigure}{0.43\textwidth}
        \centering
        \includegraphics[width=\textwidth]{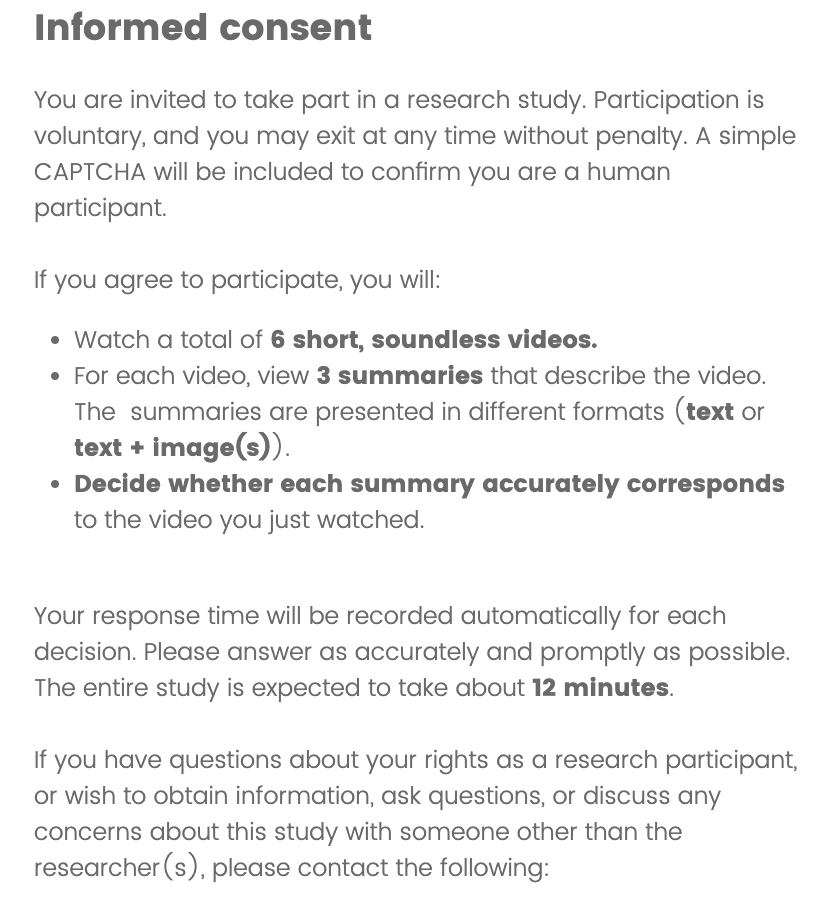}
        \caption{Instructions for correspondence test.}
    \end{subfigure}
    \caption{User Instructions for the user study.}
    \label{fig:user_ui}
\end{figure*}

All participants provide informed consent before participation and are compensated at a rate consistent with Prolific and institutional standards. 
The user instructions of the VQA and the correspondence test are presented in \cref{fig:user_ui}. 

\section{Permutation Statistical Test Results}
\label{app:permu_test}
The permutation test results (Table \ref{tab:correlations}) confirm that the information loss metric of \mmtldr maintains a statistically significant negative correlation with VQA performance across all evaluated datasets. For the MVBench and Human VQA subsets, we observe significant correlations ($p = 0.021$ and $p = 0.017$, respectively), while the LongVideoBench subset demonstrates an even stronger level of significance ($p = 0.006$). The consistent negative Pearson’s $r$ values, ranging from $-0.129$ to $-0.228$, validate the core hypothesis that lower information loss, as measured by \mmtldr, consistently corresponds to better video comprehension in both models and humans.

For both LongVideoBench and MVBench, the maximum- and minimum-scoring video samples were removed. 
Additionally, Gemini refuses to generate the 3-Image summary for a LongVideoBench video due to sensitive content after several retries, while still generating summaries in other formats of the same video.

\begin{table}[ht]
    \centering
    \caption{Permutation test results ($N_{\text{shuffles}} = 10,000$).}
    \label{tab:correlations}
    \resizebox{0.5\linewidth}{!}{%
        \begin{tabular}{l c c l}
            \toprule
            \textbf{Dataset} & \textbf{Sample Size} & \textbf{Pearson's $r$} & \textbf{$p$-value} \\
            \midrule
            MVBench & 162 & $-0.178$ & $0.021^{*}$ \\
            LongVideoBench & 458 & $-0.129$ & $0.006^{**}$ \\
            Human VQA & 110 & $-0.228$ & $0.017^{*}$ \\
            \bottomrule
            \multicolumn{4}{l}{\footnotesize * $p < 0.05$, ** $p < 0.01$.}
        \end{tabular}
    }
\end{table}

\section{Prompts Used}
\label{app:prompt}

\begin{promptbox}{Captioning Prompt}{captioning_prompt}
    You are a video content analyst. Your task is to generate a single, detailed, and objective descriptive paragraph for the provided video file.
    
    Please ensure your description faithfully includes, in chronological order:
    \begin{itemize}
        \item \textbf{Setting(s)}: Describe the environment(s) where the video takes place (e.g., indoor/outdoor, specific locations if identifiable).
        \item \textbf{Subjects and Objects}: Highlight the key people, animals, or significant objects. Include details on appearance, clothing, expressions, and notable features.
        \item \textbf{Sequence of Events}: Provide a clear, step-by-step account of the actions and events as they unfold from beginning to end.
        \item \textbf{Key Visual Details}: Note important visual information such as lighting, weather, or significant on-screen elements.
        \item \textbf{OCR (Optical Character Recognition)}: Transcribe any clearly visible and legible text seen in the video (e.g., signs, labels, graphics).
    \end{itemize}

  \textbf{Constraints}:
  \begin{itemize}
      \item Output only the descriptive paragraph—no introductions, explanations, or bullet points.
      \item Do NOT include interpretation, speculation, or information that cannot be directly observed in the video.
      \item Maintain an objective and neutral tone throughout.
  \end{itemize}
\end{promptbox}

\begin{promptbox}{Keyword masking prompt}{keyword_masking_prompt}
  Extract up to 20 keywords from the provided paragraph by selecting the most distinctive descriptors—specifically objects, motions, or events—that are directly relevant to the video content.
  
  Exclude the word ``video" as a keyword.
  
  Include each remaining keyword only once, using the original word form as they appear in the paragraph (do not apply stemming or lemmatization), and ensure all keywords are presented as a single lowercase word.
  
  Arrange the keywords sequentially as they appear in the paragraph.
  
  If fewer than 20 suitable keywords are identifiable, return only those present. Avoid duplicates.\\

  \textbf{Output Format}
  
  Output a list of words as a JSON array, for example:
  [``dog", ``jump", ``frisbee", ``park"]
\end{promptbox}

\begin{promptbox}{Keyframe selection}{keyframe_selection}
You are a video analysis API. Your task is to process the provided video and extract key moments.

Analyze the video and identify the three most significant keyframes that summarize the core action or story.
For each keyframe, output a \texttt{"timestamp"} that is as precise as possible, using the SMPTE timecode format
\texttt{"HH:MM:SS:FF"}, where \texttt{"FF"} represents the exact frame number within the second (not just to the nearest second).
This enables frame-accurate referencing.
\\

Return your response as a valid JSON array of objects. Each object must contain two keys:
\begin{itemize}
    \item \texttt{"timestamp"}: A string of the timestamp in SMPTE timecode \texttt{"HH:MM:SS:FF"} format.
    \item \texttt{"description"}: A string containing a brief, neutral description of the scene and its importance.
\end{itemize}

Do not include any text or explanation outside of the JSON array.
\\

\textbf{Example Response}
\begin{lstlisting}
[
    {
        "timestamp": "00:00:11:15",
        "description": "An intense explosion rocks an industrial structure, establishing the scene's chaotic and dangerous stakes."
    },
    {
        "timestamp": "00:00:16:03",
        "description": "A character in a tactical vest braces for impact inside an elevator, grounding the action with a human perspective."
    },
    {
        "timestamp": "00:00:20:15",
        "description": "A man in a white parka approaches a massive, high-tech vault, revealing the objective of the sequence."
    }
]
\end{lstlisting}
\end{promptbox}

\begin{promptbox}{Summary generation prompt}{summary_gen_prompt}
  You are a visual narrative analyst. You will be provided with:
  \begin{itemize}
      \item A Video File: The complete motion clip.
      \item Keyframes (Optional): A sequence of $N\le 3$ static images extracted from the video..
  \end{itemize}
  
  Your task is to write a concise, narrative summary. Use the video file as your primary source to understand the motion, transitions, and actions that happen between the static keyframes. The keyframes serve as the fixed anchor points for your narrative.

  \textbf{Guidelines}:\\
  If \textbf{keyframes are provided ($N \ge 0$)}:
  \begin{itemize}
      \item Refer to each keyframe chronologically using placeholders: [KEYFRAME1], [KEYFRAME2], ..., [KEYFRAMEn].
      \item Your text must form the narrative "glue." Describe only the essential actions leading up to, between, and following the keyframes.
      \item Do not describe the visual content of the keyframes themselves. The placeholders represent those visual moments; your role is to explain the transitions connecting them.
      \item Example structure: "The clip opens with [action] leading to [KEYFRAME1]... [action between frames]... resulting in [KEYFRAMEn]..."
  \end{itemize}

  If \textbf{no keyframes are provided ($N = 0$)}:
  \begin{itemize}
      \item Describe the essential actions, transitions, or scene changes in chronological order to form a coherent story from start to finish.
  \end{itemize}

  \textbf{Constraints}:
  \begin{itemize}
      \item The output must be a short paragraph of 2-3 sentences.
      \item Your output should be the summary text ONLY.
  \end{itemize}
\end{promptbox}

\begin{promptbox}{Distractor generation prompt}{distractor_gen_prompt}
  I will give you a correct summary of a video. Do not modify the first sentence.
  Extract the key facts (actor, action, object, location, event order).
  Then create 3 plausible distractor summaries that differ from the correct summary by exactly one or two factual changes.
  Use only these modification types: Attribute change (color, number, size, timing), Actor or location change, Event order change.
  Do not change facts related to purpose or causal information. Keep each distractor fluent, realistic, and similar in length and style to the original.
  Do not introduce impossible or absurd details. Do not repeat the original summary.
  Output only the \{total\_distractor\_num\} distractor summaries, formatted as a \{format\}.
  Do not include labels, explanations, or numbering—just the array. Do not ask any questions.
\end{promptbox}

\begin{promptbox}{VLM correspondence test prompt}{vlm_corresp_prompt}
Please watch the video carefully.
After the video, you will see several summaries. Each summary may include text, images, or both.

For each summary, decide whether it correctly corresponds to the video you just watched.
Consider the summary as a whole (text + images together). The alignment between the text and the images within a summary is not important; judge whether the overall summary matches what happened in the video.

If a summary does not correspond to the video, this may be because the keyframe image does not match the video, or because the text contradicts the video—for example, differences in order of events, colors of objects, or locations.

Please answer as accurately as possible. Also, give a confidence score between 1 and 5 where 1 is the lowest confidence and 5 is the highest confidence.
\end{promptbox}

\begin{promptbox}[promptcolor=blue]{\mmtldr score computation}{visil_score_compute}
You are asked to recover masked words that describe the content of a video.
The input varies depending on the summary modality.

\begin{itemize}
    \item \textbf{For video:} Given the video frames
    
    \item \textbf{For 3-image summary:} Given the three keyframe images [KEYFRAME1], [KEYFRAME2], and [KEYFRAME3] extracted from a video in the correct sequential order and a textual TLDR describing a video: \{\texttt{summary}\}.
    
    \item \textbf{For 1-image summary:} Given the single keyframe image [KEYFRAME1] extracted from a video and a textual TLDR describing a video: \{\texttt{summary}\}.
    
    \item \textbf{For text-only summary:} Given a textual TLDR describing a video: \{\texttt{summary}\}. 
\end{itemize}

Additionally, you are given the masked caption of the video: \{\texttt{masked\_caption}\}.

\paragraph{Task}
Guess all \texttt{[MASK]} words originally representing any words describing the video, e.g., first\_guess second\_guess. Return only the answers, without any explanation. Do not use quotes or commas; separate tokens with a single space.
\end{promptbox}

\begin{promptbox}[promptcolor=orange]{VQA evaluation}{vqa_eval}
  Given \{\texttt{input\_format}\}, answer concisely using only the provided information.
  
  Textual description of the video: \{\texttt{summary}\}
  
  Respond only with the letter corresponding to the correct option (A, B, C, or D). Do not include any other symbol or text.
  
  Question: \{\texttt{question}\}
  
  Options: \{\texttt{options}\}
\end{promptbox}

\end{document}